\def\year{2017}\relax
\documentclass[letterpaper]{article}
\usepackage{aaai17_anonymous}
\usepackage{times}
\usepackage{helvet}
\usepackage{courier}
\usepackage{color}
\usepackage{subfigure}
\usepackage{graphicx}
\usepackage{amsmath,amssymb} 
\usepackage{bm}

\begin{document}
%
\title{Learning Efficient Image Representation for Person Re-Identification}
\author{Yang Yang, Shengcai Liao, Zhen Lei, Stan Z. Li\\
        Center for Biometrics and Security Research \& National Laboratory of Pattern Recognition, \\Institute of Automation, Chinese Academy of Sciences, Beijing, China, 100190\\
}
\maketitle
\begin{abstract}
Color names based image representation is successfully used in person re-identification, due to the advantages of being compact, intuitively understandable as well as being robust to photometric variance. However, there exists the diversity between underlying distribution of color names' RGB values and that of image pixels' RGB values, which may lead to inaccuracy when directly comparing them in Euclidean space. In this paper, we propose a new method named soft Gaussian mapping (SGM) to address this problem. We model the discrepancies between color names and pixels using a Gaussian and utilize the inverse of covariance matrix to bridge the gap between them. Based on SGM, an image could be converted to several soft Gaussian maps. In each soft Gaussian map, we further seek to establish stable and robust descriptors within a local region through a max pooling operation. Then, a robust image representation based on color names is obtained by concatenating the statistical descriptors in each stripe. When labeled data are available, one discriminative subspace projection matrix is learned to build efficient representations of an image via cross-view coupling learning. Experiments on the public datasets - VIPeR, PRID450S and CUHK03, demonstrate the effectiveness of our method.
\end{abstract}

\section{Introduction}
Person re-identification is to match the persons across multiple cameras with non-overlapping views. It is challenging because the appearance of a person's surveillance images in different cameras may exhibit dramatic changes caused by illumination variation, as well as different camera views and body poses. To address it, researchers mainly focus on how to efficiently represent images that contain persons and/or how to accurately measure the similarities among them. In the former, both the robustness and distinctiveness of features are considered, while in the latter, cross-view information and the relationship among persons are analyzed.

\begin{figure}
  \centering
  \setcounter{subfigure}{0}
  \subfigure[]{
  \includegraphics[height=3.0cm]{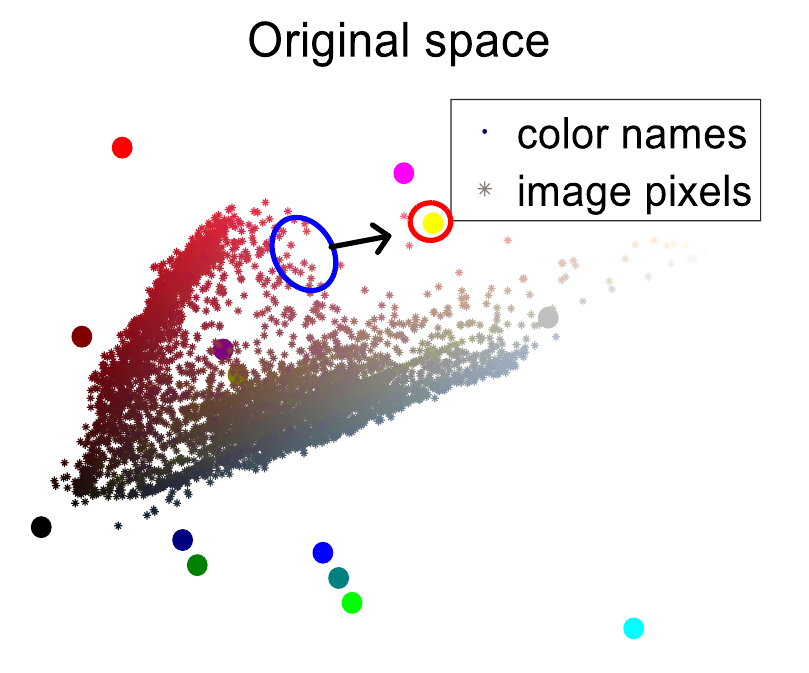}
  }
  ~~ %
  \subfigure[]{
  \includegraphics[height=3.0cm]{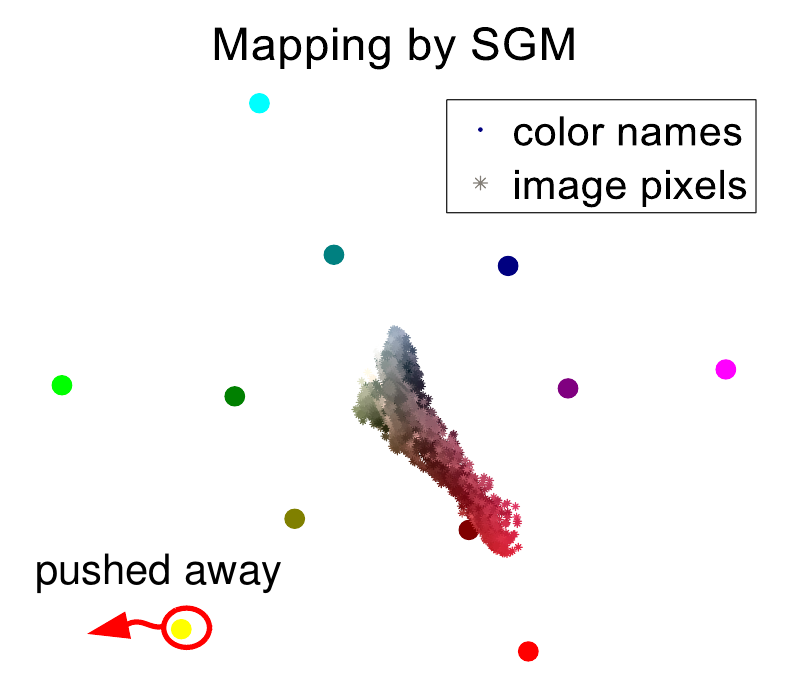}
  }
  \caption{Examples of color names and image pixels in: (a) original space and (b) transformed space by SGM. Image pixels come from an image in VIPeR dataset.}
  \label{fig: Examples of color names and image pixels (a) in original space and (b) mapped by SGM}
\end{figure}

In existing image representation methods, color features have been proved to be the most important cue in person re-identification \cite{YOYO3}. Among them, color histograms are commonly used color features \cite{R0}. However, it has the limitation of being sensitive to illumination. To increase the robustness of the color feature, a novel salient color name based color descriptor (SCNCD) is proposed in \cite{R7}. 16 pre-defined color names are successfully applied to represent person images. This kind of representation based on color names is not only compact but also has shown good robustness to illumination. The key is to develop a strategy of achieving color names descriptor (CND) to describe a pixel. Note that when it refers to a region, CND denotes a description of a region over color names (with a concatenation or pooling operation in a region). Inspired by \cite{R7}, we propose a novel CND to represent images for person re-identification.

In this paper, we argue that it is unreliable to compare image pixels with color names straightforwardly in the original Euclidean space in view of the fact that the underlying distribution of the set of color names and that of image pixels from surveillance cameras are different. This is understandable since Euclidean distance treats three color channels as an isotropic one, and thus being unable to exactly reflect the underlying relationship between color names and image pixels.  In Fig. \ref{fig: Examples of color names and image pixels (a) in original space and (b) mapped by SGM}(a), when the Euclidean distance is regarded as a dissimilarity measure between color names and image pixels, we can see that there will be a set of image pixels (circled by a blue ellipse) being assigned to the color name \emph{yellow}. In fact, the set of image pixels visually appear totally different from color name \emph{yellow}. This observation reflects that although the image pixel stays 'close' to one color name, it does not definitely imply that it has the same semantic information with the color name. The inaccurate semantic measure between image pixels and color names will further limits the performance of CND.

Motivated by Mahalanobis distance which accounts for the correlation among different dimensions, we assume a Gaussian model and propose a novel method named soft Gaussian mapping (SGM) to learn the description of an image pixel over color names. Traditional Mahalanobis distance often aggregates the color names and image pixels together and coarsely assume that they obey the same Gaussian model. However, this assumption has two deficiencies: (1) it does not take into account the difference between color names and image pixels and (2) the estimated covariance matrix may approach to reflecting the distribution of image pixels because the number of image pixels is far larger than that of color names. To overcome these problems, we model the discrepancies between color names and pixels using a Gaussian. Then, the inverse of covariance matrix are employed to bridge the gap between color names and image pixels. In Fig. \ref{fig: Examples of color names and image pixels (a) in original space and (b) mapped by SGM}(b), we can find that after the transformation in SGM, the color name \emph{yellow} has been pushed away while the image pixels and color names are similar from a semantic perspective.

To achieve the CND of a pixel, SGM further maps an image pixel to $k$ nearest (owning semantic similarity) color names. Then, an image is converted to 16 soft Gaussian maps. To establish stable and robust descriptors, a max pooling operation is imposed on the local region
\footnote{We use 3$\times$3 with a stride of 3.}
in each soft Gaussian map. With it, we can obtain a robust CND of one strip via sum pooling and sum normalization \cite{R11}. The image-level representation is finally obtained by concatenating them.

Till now, we mainly concentrate on how to learn a robust image representation. When some labeled training data are available, we further introduce cross-view coupling learning to capture the relationship among different cameras, which can be seen as an extension of XQDA \cite{R18} to two coupled variables: \emph{difference} and \emph{commonness} in \cite{YY02}. Based on it, the dimension-reduced image representations are compact and discriminative. The results of extensive experiments on three public benchmark datasets demonstrate the effectiveness of our color name based image representation and subspace learning method.

\section{Related Work}
\label{sec: Related Work}
A popular pipeline in person re-identification \cite{Y08,Y05,Y01,R24,R22,R21,R19,R8,R6,R2,YOYO15} includes image (or feature) representation and similarity learning. Among them, image representation, which is also the main concern in this paper, is arguably the most fundamental task because it determines the upper limit of the overall performance.

The appearance based low-level features can be roughly divided into (1) color (color histogram \cite{R9,R5,R0,R4,YYY01} and color names based representation \cite{R7}), (2) texture, e.g., SILTP \cite{R18,YOYO5}, LBP \cite{R0,YOYO5} and Gabor filters \cite{Y03}, (3) shape \cite{YOYO5,YYY04}, and (4) gradient\cite{YYY01}. However, due to the fact that it is extremely complicated in unconstrained surveillance condition, no single feature can be qualified completely for the task of person re-identification. Thus, a common strategy is to combine the features with complementary information to build richer signatures. Specifically, Yang et al. \cite{R7} introduce a novel salient color names based color descriptor (SCNCD) to describe colors. The extracted feature based on it has shown better performance than traditional color histogram. Since there are complementary information between them, both of SCNCD and color histogram are used to build the final holistic image representation. In addition, Cheng et al. \cite{YOYO3} propose a novel pedestrian color naming descriptor and fusing it with SCNCD, ensemble of localized features \cite{Y03} and 'mirror representation descriptors' \cite{YOYO6} to achieve better results. Three types of Local Features are combined in \cite{R4}, which models complementary aspects of human appearance: HSV histogram, Maximally Stable Color Regions and the Recurrent Highly Structured Patches. In addition, to minimize the effects of pose variations, the importance of each pixel/patch is computed according to the distance from the vertical axis, which is also used in \cite{YYY01,R7}. In \cite{R18}, an effective feature representation named local maximal occurrence (LOMO) is presented, which combines the joint HSV histogram and SILTP histogram in a three-scale pyramid model. Motivated by \cite{R18}, a novel region descriptor based on hierarchical Gaussian distribution of pixel features is proposed in \cite{YYY01}, which represents the region as a set of multiple Gaussian distributions. To represent the pixel features, the pixel location, gradient information and color information are combined together. In \cite{YOYO5}, 6 types of basic features including two types of HSV and LAB, as well as HOG and SILTP. To represent each patch, 4 different combination based on the basic features are used, each of which captures both color and texture information.

In addition to above-mentioned low-level features, attribute features which can be considered as semantic mid-level features \cite{Y05,Y12} are also introduced for person re-identification. They are often more powerful than original features in high-level tasks while owning complementary information with low-level features.

When labeled training data are available, we should learn a distinct model to make the image representation more powerful or an efficient similarity measure to calculate the similarity between any two people. Yang et al. \cite{YY01} propose a novel method to make the final image representation more discriminative -- the same persons are closer while different ones farther in the metric space. Meanwhile, to tackle the problems of different views, two coupled camera-specific dictionaries are learned in \cite{R20} and one viewpoint invariant dictionary is trained in \cite{Y02}. Recently, convolutional neural network (CNN) is adopted to learn hierarchical and discriminative image representation \cite{YYY03,DBLP:conf/icb/ZhuLYLL14,DBLP:conf/cvpr/AhmedJM15}. All of them show powerful performance when sufficient training data are available. Similarity measure learning is also another important issue for person re-identification \cite{R18,YY02,YYY03,YYY04,R25,R15,R0,DBLP:conf/icpr/YiLLL14}. Two closely related works are LSSL \cite{YY02} and XQDA \cite{R18}. Yang et al. \cite{YY02} deem that when the similarity measure method is designed, both of the \emph{difference} and \emph{commonness} of an image pair should be taken into consideration while Liao et al. \cite{R18} believe that a cross-view subspace should be learned instead of simply using PCA subspace. Our proposed cross-view coupling learning can be regarded as an extension of XQDA to the two coupled variables: \emph{difference} and \emph{commonness}. With this consideration, better results can be achieved than LSSL \cite{YY02} or XQDA \cite{R18}.

\section{Robust Image Representation}
\label{sec: Robust Image Representation}
\begin{figure*}
  \centering
  \includegraphics[width=14cm]{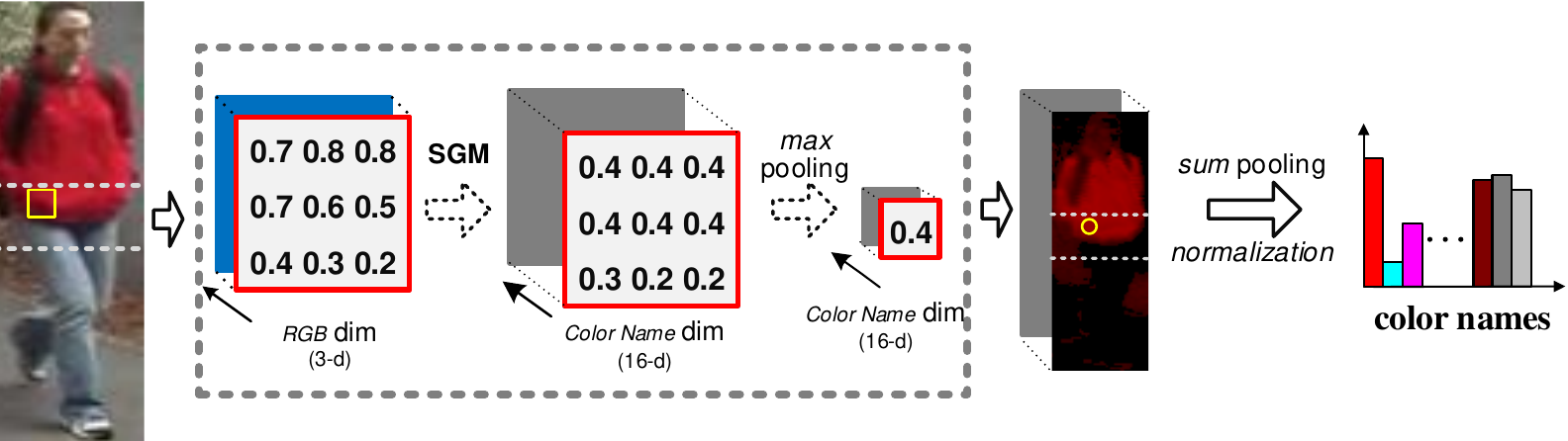}
  \caption{Flowchart of extracting a robust color names descriptor in one strip.}
  \label{fig: Flowchart}
\end{figure*}
\subsection{SGM for Color Names Descriptor of a Pixel}
\label{sec: SGM for color names descriptor of a pixel}
Let $\bm{Z}$ be a set of three-dimensional image pixels, i.e., $\bm{Z}=[\bm{z}_1,\bm{z}_2,...,\bm{z}_{n}] \in \mathcal{R}^{3 \times n}$. Meanwhile, we assume $\bm{C}=[\bm{c}_1,\bm{c}_2,...,\bm{c}_{16}] \in \mathcal{R}^{3 \times 16}$ denotes a set of 16 three-dimensional color names defined in \cite{R7}. We model the set of discrepancies between image pixels $\bm{z}_i, i=1,2,...,n$ and color names $\bm{c}_j, j=1,2,...,16$, using a Gaussian:
\begin{equation}\label{eq: Gaussian}
    P((\bm{z}_i,\bm{c}_j)|\Theta)=
              \sigma\exp(-\frac{1}{2}(\bm{z}_i-\bm{c}_j)^{T}\bm{\Sigma}^{-1}(\bm{z}_i-\bm{c}_j)),
\end{equation}
where $\sigma$ is a constant, i.e., $(2\pi)^{-3/2}|\bm{\Sigma}|^{-1/2}$ with $|\bm{\Sigma}|$ being the determinant of matrix $\bm{\Sigma}$ and $\Theta = (\bm{0},\bm{\Sigma})$ is the Gaussian model parameter. Note that the set of discrepancies is symmetric with zero mean. $\bm{\Sigma}$ is estimated by
\begin{equation}\label{eq: Estimate}
    \bm{\Sigma}= \frac{1}{16n}\sum_{i=1}^{n}\sum_{j=1}^{16}(\bm{z}_i-\bm{c}_j)(\bm{z}_i-\bm{c}_j)^T,
\end{equation}
where $\bm{\Sigma}$ is a $3 \times 3$ symmetric matrix. To further ensure that $\bm{\Sigma}$ is reversible and prevent $\bm{\Sigma}^{-1}$ from being close to non-positive one, we make an eigenvalue rectification by tuning up the non-positive eigenvalues of $\bm{\Sigma}$
\footnote{In most cases, this step can be neglected in consideration that $n\gg3$. However, it should be done when we can only obtain images with poor quality, e.g., many dark clothing with low-resolution or captured with insufficient light.}
. Based on eigenvalue decomposition, we have $\bm{\Sigma}= \sum_{i=1}^{3}\lambda_{i} \bm{u}_{i}\bm{u}_{i}^{T}, i=1,2,3$. $\lambda_{i}$ is eigenvalue and $\bm{u}_i$ is corresponding eigenvector. The rectified covariance matrix is $\widetilde{\bm{\Sigma}} = \sum_{i=1}^{3}f(\lambda_{i},\epsilon_0) \bm{u}_{i}\bm{u}_{i}^{T}$ ($\epsilon_0$ is small value, we set it to $1e^{-4}$ throughout the paper) with
\begin{equation}\label{eq: Rectification}
f(\lambda_{i},\epsilon_0) =
 \left\{
 \begin{array}{l@{\quad,\quad}l}
   \lambda_{i} & \lambda_{i}>0\\
   \epsilon_0^{-1} & \lambda_{i} \leqslant 0
 \end{array}.
 \right.
\end{equation}
Then, the inverse of rectified covariance matrix can be computed as
\begin{equation}\label{eq: Sigma}
\widetilde{\bm{\Sigma}}^{-1} = \sum_{i=1}^{3}f(\lambda_{i},\epsilon_0)^{-1} \bm{u}_{i}\bm{u}_{i}^{T}.
\end{equation}

With Eqs. \ref{eq: Gaussian} and \ref{eq: Sigma}, we can easily estimate the likelihood of the image pixel $\bm{z}_i$ belonging to the color name $\bm{c}_i$. This estimated likelihood can be served as the color names descriptor of an image pixel. It describes the membership of an image pixel to color names from a probabilistic perspective. In \cite{R7,R11,R12,YOYO7}, an 'early cut-off' is often used to remove the adverse impact of dissimilar factor and can handle the underlying manifold structure when local descriptors are learned. However, Yang et al. \cite{YY01} also point out that when image-level descriptors are learned, the strategy of 'early cut-off' may harm the performance because with only several nonzero value, the obtained descriptors can not
retain its original data information. To make it more generic, we defined our soft gaussian mapping in a more flexible manner:
\begin{equation}\label{eq: SGM}
\bm{s}_{j} =
 \left\{
 \begin{array}{l@{\quad,\quad}l}
   P((\bm{z}_i,\bm{c}_j)|\Theta) & if ~\bm{c}_j \in \mathcal{N}_k(\bm{z}_i)\\
   0 & else,
 \end{array}
 \right.
\end{equation}
where $\mathcal{N}_k(\bm{z}_i)$ denotes $k$ most similar color names of $\bm{z}_i$ defined by their similarities $P((\bm{z}_i,\bm{c}_j)|\Theta)$. We further employ sum normalization \cite{R11}
\begin{equation}\label{eq: sum normalization}
\bm{s}^T1=1
\end{equation}
to make the descriptor stable. In consequence, an image pixel's CND obtained by SGM can be taken as its probability distribution over color names $\bm{s}_{j}, j=1,2,...,16$. Here, 'soft' means that given an image pixel, several color names are considered. For the task of pedestrian color naming \cite{YOYO3}, we set $k$ to 1. When local descriptor is learned, it is set as a small number (around 5). When it is used to learn higher-level representation, it is optimum to use all of the color names.\\

\noindent\textbf{Remark} Under the Gaussian model of the discrepancies, we can employ the inverse of matrix, i.e., $\widetilde{\bm{\Sigma}}^{-1}$ in Eq. \ref{eq: Sigma} to bridge the distribution gap between the color names and image pixels while measuring their similarity in an accurate manner. This is reflected in Fig. \ref{fig: Examples of color names and image pixels (a) in original space and (b) mapped by SGM} which compares the original Euclidean distance and mapping space. Note that the mapping space (or transform matrix) by SGM are obtained using the Cholesky decomposition of $\widetilde{\bm{\Sigma}}^{-1}$. Then, we explain it through an analysis of $\bm{\Sigma}$. We rewrite Eq. \ref{eq: Estimate} as
\begin{equation}\label{eq: Analysis}
    \bm{\Sigma}= \frac{1}{16n}\sum_{i=1}^{n}\sum_{j=1}^{16}((\bm{z}_i\bm{z}_i^T+\bm{c}_j\bm{c}_j^T)-(\bm{z}_i\bm{c}_j^T+\bm{c}_j\bm{z}_i^T)),
\end{equation}
where the first two terms denote the overall Gaussian distribution while the last two terms reflect the correlation between the color names and image pixels. By the subtracting the correlation from the overall distribution, the obtained covariance matrix in Eq. \ref{eq: Analysis} only manifests the discrepancy. This discrepancy are further removed by using the inverse of covariance matrix, i.e., $\bm{\Sigma}^{-1}$(or $\widetilde{\bm{\Sigma}}^{-1}$). This is why we can accurately measure them in the transformed space.

\subsection{Image Representation: from Pixel-Level to Image-Level }
\label{sec: Image Representation}
Based on SGM, an image can be converted to 16 soft Gaussian maps. Then, in each map, we are required to characterize the signature of each stripe (using 10 non-overlapping horizontal stripes). In \cite{R7}, average pooling is directly used in each stripe. However, in consideration that different camera view and poses may lead to a variance even in a small enough local patch, we then first focus on how to describe the statistical information in a local patch.

When the local patch is small enough ($\leqslant 3\times3$), the descriptors should be the similar in each stripe. Thus, we take the maximum in the local patch as its descriptor of a patch. By doing so, small deviation or mapping noises can be removed. As shown in Fig. \ref{fig: Flowchart}, after SGM, the probability values in the $3 \times 3$ patch of the first map are various. We then perform max pooling in the local patch and only 0.4 is taken as the descriptor of this patch, regardless of other values. Sum pooling \cite{R11} and sum normalization in Eq. \ref{eq: sum normalization} are then employed to achieve the robust CND of a stripe. The flowchart of extracting a robust color names descriptor in one strip is shown in Fig. \ref{fig: Flowchart}. Finally, the robust color names based image representation is obtained via concatenating them.

\section{Cross-View Coupling Learning}
\label{sec: Cross-View Coupling Learning}
Liao et al. \cite{R18} propose a method XQDA to reduce the dimension of the original features which is often large. A discriminative subspace is learned via maximizing
\begin{equation}\label{eq: subspace learning}
J(\bm{w}) = \sigma_E(\bm{w})/\sigma_I(\bm{w}),
\end{equation}
where $\sigma_E(\bm{w})$ and $\sigma_I(\bm{w})$ denotes inter-personal and intra-personal variances, respectively.
From Eq \ref{eq: subspace learning}, we can find that in the learned subspace, the intra-personal variance is suppressed with respect to inter-personal variance. In addition, Yang et al. \cite{YY02} point out that when both \emph{commonness} and \emph{difference} are taken into consideration, more distinctiveness can be expected. Then, we propose a new subspace learning method which can be considered as an extension version of the XQDA (only considers \emph{difference}) to both \emph{commonness} and \emph{difference}.

Given an image pair ($\bm{x}, \bm{y}$) from two different cameras, \emph{commonness} $\bm{m}$ and \emph{difference} $\bm{e}$ are defined as
\begin{equation}\label{eq: E and M}
\left\{
\begin{array}{rl}
\bm{m} = \bm{x} + \bm{y}\\
\bm{e} = \bm{x} - \bm{y}
\end{array}
\right.
\end{equation}
According to \cite{YY02}, the similarity score between two people can then be computed via
\begin{equation}\label{eq: similarity measure}
r(\bm{x},\bm{y}) = \bm{m}^{T}(\bm{\Sigma}^{-1}-\bm{\Sigma}_{m\mathcal{I}}^{-1})\bm{m}
          -\bm{e}^T(\bm{\Sigma}_{e\mathcal{I}}^{-1}~-\bm{\Sigma}^{-1})\bm{e}
\end{equation}
with
\begin{equation}\label{eq: me sigma}
\bm{\Sigma} = \frac{1}{2}(\bm{\Sigma}_{m\mathcal{I}} + \bm{\Sigma}_{e\mathcal{I}}).
\end{equation}
where the inter-personal covariance matrix of $\bm{m}$ is equivalent to that of $\bm{e}$, each of them is computed by $\bm{\Sigma}$.

In Eq. \ref{eq: E and M}, the defined coupled variables $\bm{m}$ and $\bm{e}$ (zero-centered) are negative correlation, i.e., when $\|\bm{m}\|_2$ is small, $\|\bm{e}\|_2$ is large and vice versa. Then, we expect to learn a subspace owning the following traits: (1) for $\bm{e}$, the intra-personal variance is suppressed with respect to inter-personal variance (consistent with XQDA) and (2) for $\bm{m}$, the inter-personal variance is suppressed with respect to intra-personal variance. To that end, our objective is defined to maximize $J_e(\bm{w})$ and $J_m(\bm{w})$ jointly:
\begin{equation}\label{eq: CCP0}
\left\{
\begin{array}{ll}
J_e(\bm{w}) = \sigma_{eE}(\bm{w})/\sigma_{eI}(\bm{w})\\
J_m(\bm{w}) = \sigma_{mI}(\bm{w})/\sigma_{mE}(\bm{w})
\end{array}
\right.
\end{equation}
where $\sigma_{eE}$ and $\sigma_{mE}$ denote the inter-personal variance covariance matrices over $\bm{e}$ and $\bm{m}$, respectively while $\sigma_{eI}$ and $\sigma_{mI}$ denote the intra-personal variance covariance matrices over $\bm{e}$ and $\bm{m}$, respectively.
According to \ref{eq: me sigma}, $\sigma_{eE}$ equals to $\sigma_{mE}$ in the same subspace. Consequently, our objective can be simplified to maximize $J_0(\bm{w})$
\begin{equation}\label{eq: CCP1}
J_0(\bm{w}) = \sigma_{mI}(\bm{w})/\sigma_{eI}(\bm{w}),
\end{equation}

Eq. \ref{eq: CCP1} reflects that when only similar pairs are considered, we wish to learn a subspace where the intra-personal variance for $\bm{e}$ is suppressed with respect to intra-personal variance for $\bm{m}$ in the learned subspace.
As in \cite{R18}, we rewrite Eq. \ref{eq: CCP1} to
\begin{equation}\label{eq: CCP2}
J_0(\bm{w}) = \bm{w}^T\Sigma_{mI}\bm{w}/\bm{w}^T\Sigma_{eI}\bm{w},
\end{equation}
Similar to LDA, the maximization of $J_0(\bm{w})$ can be solved by generalized eigenvalue decomposition problem, i.e., the subspace is composed of the eigenvectors corresponding to $r$ largest eigenvalue of $\Sigma_{eI}^{-1}\Sigma_{mI}$ (refer to \cite{R18} for detail). Since the cross-view information is learned based on two coupled variables, we name our subspace learning method as cross-view coupling learning (CCL).

Once the subspace is learned from labeled training data (based on \ref{eq: CCP2}), we employ it to reduce the dimension of the image-level representation. Finally, we employ Eq. \ref{eq: similarity measure} to calculate the similarity score for any two images.

\section{Experiments}
\label{sec: Experiments}
In this section, we evaluate our method on three publicly available datasets (VIPeR dataset \cite{R14}, PRID 450S dataset~\cite{R15} and CUHK 03 dataset~\cite{YOYO16}). The best matching rate is shown in red while our methods are shown in bold. It is evaluated on a PC with the 3.40 GHz Core I7 CPU with 8 cores.
\subsection{Datasets}
\noindent\textbf{VIPeR Dataset.}
VIPeR dataset has 632 persons captured with two disjoint cameras in outdoor environments. There is one image for each person in each camera view. It is challenging due to arbitrary viewpoints, pose changes and illumination variations. Images are mostly captured from 0 degree to 90 degree in Camera A while those from Camera B mostly from 90 degree to 180 degree. All images are normalized to 128$\times$48.

\noindent\textbf{PRID450S Dataset.}
PRID450S dataset consists of 450 persons captured from two spatially disjoint camera views. Each person has one image in each view. Due to different viewpoint changes, background interference, partial occlusion and illumination variations, it is also a challenging dataset. All images are normalized to 168$\times$80.

\noindent\textbf{CUHK03 Dataset.}
CUHK03 dataset contains 1360 persons captured with six surveillance cameras. Each person is observed by two disjoint camera views and has an average of 4.8 images in each view. It is also a challenging dataset because there are occlusions, misalignment and body part missing. All images are normalized to 160$\times$60 pixels.

\subsection{Setup}
\label{sec: Setup}
\noindent\textbf{Training/test.}
In experiments, we report our results in form of Cumulated Matching Characteristic (CMC) curve~\cite{R16}. For VIPeR and PRID450S datasets, we randomly chose 50\% of all persons training while the remaining are used for test. We conduct the evaluation procedure for 10 random splits and adopt the single-shot scheme. For CUHK03 dataset, 1160 persons are employed for training while 100 persons for test. The experiments are conducted with 20 random splits and the multi-shot evaluation strategy is employed.

\noindent\textbf{Features.}
As in \cite{R7}, we use the image-foreground representation
\footnote{On VIPeR and PRID450S datasets, we use the same mask as \cite{R7}. On CUHK03 dataset, we use the method in~\cite{Y09} to automatically generate the masks (0.13s for an image of 160$\times$60).}
and the same 4 color spaces including RGB, rgb, $l_1l_2l_3$ and HSV. However, as is discussed in Sec. \ref{sec: Related Work}, no single feature can be qualified for the unconstrained surveillance condition. Considering that color names based representation has complementary information with color histogram and texture, we also employ another two simple features to achieve a new state-of-the-art results: color histogram and SILTP. We regard them as three types of features: color names descriptors, color histogram and SILTP.

\noindent\textbf{Parameters.}
Unless otherwise specified, we empirically set the parameters as follows: (1) We employ 10 non-overlapping stripes. (2) We set $k$ to 5 for SGM. (3) We adopt 16 bins in each channel for color histogram. (4) We project each type of features to 100 subspace by CCL.

\subsection{Evaluation on VIPeR}
\label{sec: Evaluation on VIPeR}
In this subsection, we make a thorough evaluation of our method on the widely used VIPeR dataset.

\noindent\textbf{Different Features.}
We simply name our color names based image representation as SGM. We compare SGM with SGM(Eu) and SGM(Ma). SGM(Eu) means that under our framework, the covariance matrix $\bm{\Sigma}$ in Eq. \ref{eq: Gaussian} is set to be an identity matrix. SGM(Ma) denotes that the covariance matrix $\bm{\Sigma}$ in Eq. \ref{eq: Gaussian} is computed based on a set of pixels aggregated together. We further make a comparison with SCNCD \cite{R7} which is based on an index table of color names. In addition, we also compare SGM with the used color histogram (CH) and SILTP.
\begin{table}
\begin{center}
\begin{tabular}{|l||c|c|c|}
\hline
\bf Method           &\bf Rank 1     &\bf Dim &\bf Time (s)
\\ \hline  \hline
SCNCD       &44.6\%        &1280   & 0.030\\
CH          &37.8\%        &3480   & 0.018\\
SILTP       &18.9\%        &6480   & 0.004\\
\hline  \hline
\textbf{SGM(Eu)}     &\textbf{41.6\%} &\textbf{1280}&\textbf{0.033}\\
\textbf{SGM(Ma)}     &\textbf{44.9\%} &\textbf{1280}&\textbf{0.036}\\
\textbf{SGM}         &\color{red}\textbf{50.0\%} &\textbf{1280}&\textbf{0.036}\\
\hline
\end{tabular}
\end{center}
\caption{Comparison with different features on VIPeR dataset.}
\label{Table: Comparison with different features on VIPeR dataset}
\end{table}
In Table \ref{Table: Comparison with different features on VIPeR dataset}, we can find that based on CCL, our proposed SGM performs best and improves the performance of SCNCD by 5.4\%. In addition, color features perform better than texture feature, which shows the importance of color features in person re-identification. Among those color names based image representations, SGM(Eu) performs the worst. This demonstrates that when the color names are directly compared with image pixels in the original Euclidean space, the different underlying distributions may cause inaccurate semantic measure, which will further limits the final performance. SGM(Ma) aggregates the image pixels and color names together and the learned covariance matrix is used to eliminate the inaccurate representation to a certain degree. In comparison with SGM(Eu), SGM(Ma) improves the results by 3.3\% and slightly better than SCNCD. By eliminating the discrepancy, SGM performs best. The extracting time based on SGM is also competitive: 0.036s for a 128$\times$48 image.

\noindent\textbf{Evaluation of Learning based Methods.}
\begin{table}
\begin{center}
\begin{tabular}{|l||c|c|c|c|}
\hline
\bf{Dimension}  &\bf 60  &\bf 80   &\bf 100  &\bf 200
\\ \hline \hline
LSSL       &45.5\%   &46.0\%  &47.3\%  & 47.0\%   \\
XQDA       &43.1\%   &44.4\%  &44.4\%  & 44.4\%   \\
KISSME     &38.6\%   &40.7\%  &40.9\%  & 33.8\%   \\
\hline  \hline
\textbf{CCL}          &\color{red}\textbf{49.0\%}&\color{red}\textbf{49.6\%}&\color{red}\textbf{50.0\%}
                      &\color{red}\textbf{50.0\%}\\
\hline
\end{tabular}
\end{center}
\caption{Evaluation of Learning based Methods on VIPeR. Rank 1 results are shown with different dimensions.}
\label{Table: Evaluation of Learning based Methods}
\end{table}
In Table \ref{Table: Evaluation of Learning based Methods}, we compare the proposed CCL with other closely related learning methods: KISSME \cite{R0}, XQDA \cite{R18} and LSSL \cite{YY02}. For a fair comparison, all of them are based on 1280-dimensional SGM and Rank 1 results are shown with different dimensions: 60, 80, 100 and 200. We can observe that CCL consistently performs better than compared methods.

\subsection{Comparison with the State-of-the-art Methods}
\label{sec: Comparison with the State-of-the-art Methods}
We compare our methods with the state-of-the-art approaches (in recent four years) on VIPeR, PRID450S and CUHK03 datasets, respectively. Our methods consist two parts: (1) we only use color names based representation (named SGM) and (2) we fuse \textbf{S}GM, \textbf{c}olor histograms and \textbf{S}ILTP (named SCS).

\noindent\textbf{State-of-the-art: VIPeR.}
VIPeR is a classic benchmark dataset and may be most widely used for person re-identification. Due to arbitrary viewpoints, pose changes and illumination variations, it is very challenging and remains unsolved. The compared approaches include SCSP+PCN \cite{YOYO3}, SCSP \cite{YOYO5}, WLC \cite{YOYO8}, GOG \cite{YYY01}, LSSL \cite{YY02}, MetricEn \cite{YYY04}, LOMO \cite{R18} and SCNCD+CH \cite{R7}.
\begin{table}
\begin{center}
\begin{tabular}{|l||c|c|c|c|}
\hline
\bf Rank              &\bf 1             &\bf 5             &\bf 10            & 20
\\ \hline \hline
SCSP+PCN  &54.2\%            &82.8\%            &91.4\%            & \color{red}\textbf{99.1}\%\\
SCSP      &53.5\%            &82.6\%            &91.5\%            & 96.7\%\\
WLC       &51.4\%            &76.4\%            &84.8\%            &  -\\
GOG       &49.7\%            &79.7\%            &88.7\%            & 94.5\%\\
LSSL      &47.8\%            &77.9\%            &87.6\%            & 94.2\%\\
MetricEn  &45.9\%            &77.5\%            &88.9\%            & 95.8\%\\
LOMO      &40.0\%            &68.1\%            &80.5\%            & 91.1\%\\
SCNCD+CH  &37.8\%            &68.5\%            &81.2\%            & 90.4\%\\
\hline   \hline
\textbf{SGM}          &\textbf{50.0\%}   &\textbf{78.5\%}   &\textbf{88.1\%}     &\textbf{94.2\%}\\
\textbf{SCS}          &\color{red}\textbf{59.2\%}&\color{red}\textbf{85.0\%}&\color{red}\textbf{92.1\%}
                      &\textbf{96.4\%}\\
\hline
\end{tabular}
\end{center}
\caption{Comparison with the state-of-the-art methods (in recent three years) on VIPeR dataset.}
\label{Table: Results-VIPeR}
\end{table}
Among the previous approaches in Table \ref{Table: Results-VIPeR}, SCSP+PCN achieves the best result at Rank 1. It combines the color names descriptor (PCN) with SCSP which uses 6 types of basic features including two types of HSV and LAB, as well as HOG and SILTP. In comparison, our SGM with only color information is also promising. When it is combined with CH and SILTP, we can achieve the best result 59.2\% at Rank 1 (5.0\% higher than SCSP+PCN \cite{YOYO3}).

\noindent\textbf{State-of-the-art: PRID450S.}
\begin{table}
\begin{center}
\begin{tabular}{|l||c|c|c|c|}
\hline
{\bf Rank}           &\bf 1   &\bf 5   &\bf 10   &\bf 20
\\ \hline \hline
GOG       & 68.4\% & 88.8\% & 94.5\%  & \color{red}97.8\% \\
LOMO      & 62.6\% & 85.6\% & 92.0\%  & 96.6\% \\
MED$\_$VL & 45.9\% & 73.0\% & 82.9\%  & 91.1\% \\
TSR       & 44.9\% & 71.7\% & 77.5\%  & 86.7\% \\
CSL       & 44.4\% & 71.6\% & 82.2\%  & 89.8\% \\
SCNCD+CH  & 41.6\% & 68.9\% & 79.4\%  & 87.8\% \\
\hline  \hline
\textbf{SGM}          &\textbf{66.1\%}   &\textbf{86.9\%}   &\textbf{91.4\%}     &\textbf{95.3\%}\\
\textbf{SCS}          &\color{red}\textbf{74.8\%} &\color{red}\textbf{91.4\%} &\color{red}\textbf{94.8\%}
                      &\textbf{97.2\%}\\
\hline
\end{tabular}
\end{center}
\caption{Comparison with the state-of-the-art methods (in recent four years) on PRID450S dataset.}
\label{Table: Results-PRID 450S}
\end{table}
we compare our method with the state-of-the-art approaches on PRID450S dataset, including GOG \cite{YYY01}, LOMO \cite{R18}, MED$\_$VL \cite{YY01}, CSL \cite{Y01}, TSR \cite{Y12} and SCNCD+CH \cite{R7}.

Among the previous approaches in Table \ref{Table: Results-PRID 450S}, GOG achieves the best results at Ranks 1-20. The second best results are achieved by LOMO \cite{R18}. Both of them use color and texture information. Our SGM with only color information performs better than MED$\_$VL and SCNCD+CH\cite{R7}, both of which are also based on color features. Our SCS which combines color and texture information achieves a new state-of-the-art result (74.8\%) at Rank 1.

\noindent\textbf{State-of-the-art: CUHK03.}
We also test our method on the large dataset CUHK03 with both labeled and detected setting. The compared methods include JLML\cite{YOYO17}, GOG \cite{YYY01}, DNS \cite{YOYO9}, MetricEmsemble \cite{YYY04}, and LOMO \cite{R18}. Among the previous approaches in Table \ref{Table: Results-CUHK 03}, JLML\cite{YOYO17} achieves the best results. It is based on the ResNet. On large datasets, the learned CNN features outperform those learned by traditional methods by a large margin. Compared with traditional methods, we can still achieve promising results (at Rank 1): 79.6\% and 76.7\% with labeled and the automatically detected bounding boxes, respectively.

\begin{table}
\begin{center}
\begin{tabular}{|l||c|c|}
\hline
      &\bf Labeled      &\bf Detected
\\ \hline \hline
JLML            & \color{red}\textbf{83.2}\%          & \color{red}\textbf{80.6}\%      \\
GOG             & 67.3\%          & 65.5\%      \\
DNS             & 62.6\%          & 54.7\%      \\
MetricEmsemble  & 62.1\%          & -         \\
LOMO            & 52.2\%          & 46.3\%      \\
\hline  \hline
\textbf{SCS}    & \textbf{79.6}\%          & \textbf{76.7}\%      \\
\hline
\end{tabular}
\end{center}
\caption{Comparison with the state-of-the-art methods (in recent four years) on CUHK03 dataset. Rank 1 results are shown with both labeled and detected setting.}
\label{Table: Results-CUHK 03}
\end{table}


\section{Conclusion}
\label{sec:Conclusion}
In this paper, we propose a new method to learn color names based image representation. It addresses the semantic gap between color names and image pixels based on a Gaussian model and improves the performances of existing color features based approaches for person re-identification. To make the representation more discriminative, a new subspace learning method is presented by a cross-view analysis on two coupled variables: \emph{commonness} and \emph{difference}. We make a thorough evaluation of our method on VIPeR and also demonstrate its effectiveness by comparing with state-of-the-art approaches on three public datasets.

{\small
\bibliographystyle{aaai}
\bibliography{egbib}
}

\end{document}